\documentclass[11pt,a4paper]{article}
\usepackage[hyperref]{eacl2021}
\usepackage{times}
\usepackage{latexsym}
\usepackage{amssymb}
\newdimen\hfuzz

\usepackage{mathtools}
\usepackage{microtype}
\usepackage{graphicx}
\usepackage{amsmath}
\aclfinalcopy 
\newtheorem{definition}{Definition}

\title{Low-Resource Neural Machine Translation for Southern African Languages}

\author{Evander Nyoni $^a$ \and Bruce A. Bassett$^{a,b,c}$\\
  $^a$African Institute for Mathematical Sciences, 6 Melrose Road, Muizenberg, 7945, South Africa \\
  $^b$ Department of Maths and Applied Maths, University of Cape Town, Cape Town, South Africa \\
  $^c$ South African Astronomical Observatory, Observatory, Cape Town, 7925, South Africa }

\date{}

\begin{document}
\maketitle
\begin{abstract}
Low-resource African languages have not fully benefited from the progress in neural machine translation because of a lack of data. Motivated by this challenge we compare zero-shot learning, transfer learning and multilingual learning on three Bantu languages (Shona, isiXhosa and isiZulu) and English. Our main target is English-to-isiZulu translation for which we have just 30,000 sentence pairs,  $28\%$ of the average size of our other corpora. We show the importance of language similarity on the performance of English-to-isiZulu transfer learning based on English-to-isiXhosa and English-to-Shona parent models whose BLEU scores differ by $5.2$. We then demonstrate that multilingual learning  surpasses both transfer learning and zero-shot learning on our dataset, with BLEU score  improvements relative to the baseline English-to-isiZulu model of $9.9$, $6.1$ and $2.0$ respectively. Our best model also improves the previous SOTA BLEU score by more than $10$.
\end{abstract}

\section{Introduction}
\begin{figure*}[!htbp]
    \centering
    \includegraphics[width=\textwidth, height=5cm]{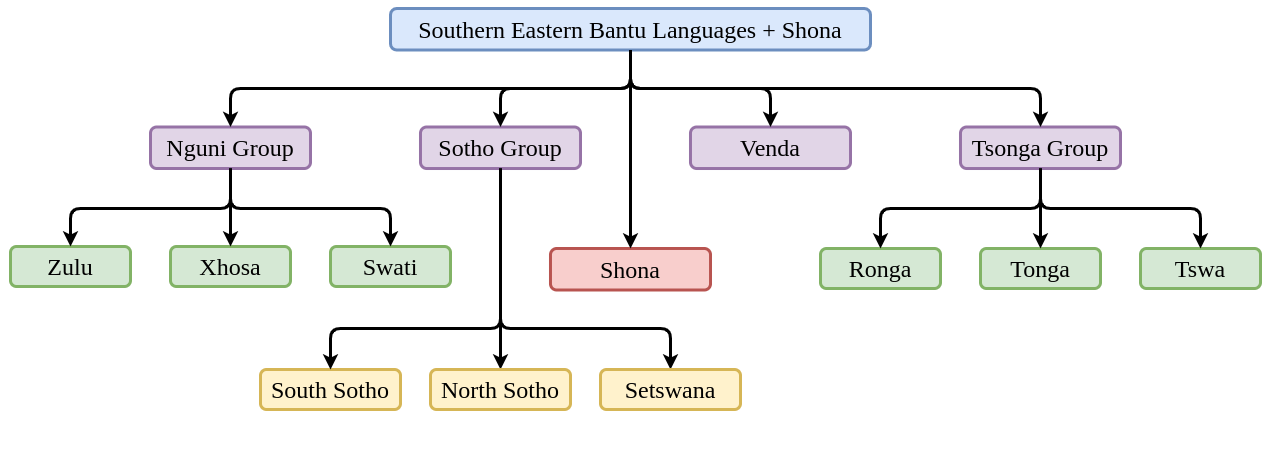}
    \caption{The language family tree of the South-Eastern Bantu languages including Shona. This tree demonstrates the relationship between languages with languages of the same subclass being closely related \cite{doke2014english}.}
    \label{fig: lang_fam_tree}
\end{figure*}

Machine translation (MT) is the automation of text (or speech) translations from one language to another, by a software. The rise of MT commenced sometime in the mid-twentieth century \cite{weaver1955translation} and has since the early 1900s experienced rapid growth due to the increase in the availability of parallel corpora and computational power \cite{marino2006n}. Subject to the inception of neural machine translation (NMT) \citep{sutskever2014sequence, bahdanau2014neural}, MT has seen substantial advancements in translation quality as modern MT systems draw closer and closer to human translation quality. Notwithstanding the progress achieved in the domain of MT, the idea of NMT system development being data-hungry remains a significant challenge in expanding this work to low resource languages.\\ \indent Unfortunately, most African languages fall under the low-resourced group and as a result, MT of these languages has seen little progress. Irrespective the great strategies (or techniques) that have been developed to help alleviate the low resource problem, African languages still have not seen any substantial research or application of these strategies \cite{orife2020masakhane}. With a considerable number of African languages being endangered \cite{ethnologue}, this shows that these languages are in dire need of MT translation tools to help save them from disappearing. In other words, this poses a challenge to the African community of  NLP practitioners.\\ \indent This paper examines the application of low-resource learning techniques on African languages of the Bantu family. Of the three Bantu languages under consideration, isiZulu, isiXhosa and Shona, the first two fall under the Nguni language sub-class indicating a close relationship between the two, as shown in Figure \ref{fig: lang_fam_tree}. Shona is not closely related to the Nguni language sub-class \cite{holden2003spread}. Comparing MT on these three gives us the opportunity to explore the effect of correlations and similarities between languages. We give a comparative analysis of three learning protocols, namely, \emph{transfer learning}, \emph{zero-shot learning} and \emph{multilingual modeling}.\\ \indent Our experiments indicate the tremendous opportunity of leveraging the inter-relations in the Bantu language sub-classes, to build translation systems. We show that with the availability of data, multilingual modeling provides significant improvements on baseline models. In the case of transfer learning, we show that language sub-class inter-relations play a major role in translation quality. Furthermore, we demonstrate that zero-shot learning only manages to surpass transfer learning when we employ a parent model that is distantly related to the task of interest.

\indent The following sections of this paper are presented as follows: in Section \ref{sec: background} we briefly review NMT and the main architecture we use in this work. Section \ref{sec: training_protocols} then gives a brief outline of the training protocols employed in this paper. Thereafter, we discuss the work that is related to this paper in Section \ref{sec: related_work}. Section \ref{sec: experiments} discusses our experiments on the translation of English-to-Shona (E-S), English-to-isiXhosa (E-X), English-to-isiZulu (E-Z) and isiXhosa-to-isiZulu (X-Z). Finally, the results and conclusion of this work are outlined in Sections \ref{sec: results} and \ref{sec: conclusions} respectively.

\section{Background}\label{sec: background}
Modern NMT models have the encoder-decoder mechanism \cite{sutskever2014sequence} as the vanilla architecture. Given an input sequence $x=(x_{1},...,x_{S})$ and target sequence $y=(y_{1},...,y_{T})$. A NMT model decomposes the conditional probability distribution over the output sentences as follows: 
\begin{equation}
\begin{split}
    p(y|x) &= \prod^{T}_{t=1} p(y_{t}|y_{1:t-1},x_{1:S})
\end{split}
\end{equation}
Jointly trained, the encoder-decoder mechanism learns to maximize the conditional log-likelihood
\begin{equation}\label{eq: enc_dec_logL}
    \max_{\theta}\;\frac{1}{|x|}\sum_{(x,y)\in D} \log p_{\theta}(y|x),
\end{equation}
where $D$ denotes the set of training pairs and $\theta$ the set of parameters to be learned.

Several encoder-decoder architectures have been developed and they each model the probability distribution $P(y_{i}|y_{1:i-1}, x)$ differently. For example, appertaining to the general encoder-decoder architecture by \citet{sutskever2014sequence}, the introduction of an encoder-decoder mechanism with attention \cite{bahdanau2014neural} has produced significant improvements on translation quality. The \citet{bahdanau2014neural} architecture modifies the factorized conditional probability as follows:
\begin{equation}
\begin{split}
    p(y|x) &= \prod^{T}_{t=1} p(y_{t}|y_{1:t-1},x_{1:S}) \\
    &= \prod^{T}_{t=1}g(y_{t}|y_{1:t-1},s_{t},c(x))
\end{split}
\end{equation}

where $g(\cdot)$ denotes the decoder which generates the target output in auto-regressive fashion, and $s_{t}$ denotes decoder recurrent neural network (RNN) hidden state. The widely used RNN variants are the gated recurrent units (GRU) and the long term short term memory (LSTM) networks as these help mitigate the vanishing gradient problem \cite{pouget2014overcoming}. The encoder compresses the input sequence into a fixed length vector $c(x)$, which is widely known as the context vector. This vector is then relayed to the decoder, where it is taken in as input.

Attention-based models \citep{ bahdanau2014neural,chorowski2015attention, luong2015effective} have become the conventional architectures of most sequence to sequence transduction tasks. In the domain of NMT, the transformer model introduced by \citet{vaswani2017attention} has become the vanilla architecture of NMT and has produced remarkable results on several translation tasks. In this work, we employ the transformer model in all our training protocols. The transformer architecture's fundamental unit is the multi-head attention mechanism that is solely based on learning different sequence representations with its multiple attention heads.  

The key aspect of the transformer architecture is its ability to jointly pay attention to or learn different sequence representations that belong to different sub-spaces of distinct positions. In other words, the transformer model focuses on learning pair relation representations which in turn are employed to learn the relations amongst each other. The multi-head attention mechanism consists of $N$ attention layers such that for each head, the query and key-value pairs are first projected to some subspace with some dense layers of sizes $h_{q}$, $h_{k}$, and $h_{v}$ respectively. Suppose the query, key and value have dimensions $n_{q}$, $n_{k}$ and $n_{v}$ respectively, each attention head then maps the query and key-value pairs to some output $o^{(j)}$ as follows
\\
\begin{equation}
    o^{(j)} = \sigma(W^{(j)}_{q}q, W^{(j)}_{k}k, W^{(j)}_{v}v)
\end{equation}
where $\sigma $ is a dot-product attention function, $W^{(j)}_{q} \in \mathbb{R}^{h_{q} \times n_{q}}$, $W^{(j)}_{k} \in \mathbb{R}^{h_{k} \times n_{k}} $ and $W^{(j)}_{v} \in \mathbb{R}^{h_{v} \times n_{v}} $. Thereafter, the $N$ outputs of length $h_{q} = h_{v} = h_{k}$  each, are then concatenated and fed to a dense layer of $n_{o}$ hidden units, to produce a multi-head attention output
\begin{equation}
    o = W_{o}\begin{bmatrix}
o^{(1)}\\
\vdots\\
o^{(N)}\\
\end{bmatrix}
\end{equation}
where $W_{o} \in \mathbb{R}^{n_{o} \times Nh_{v}} $. In all our training protocols (or experiments), we employ the transformer architecture with 6 encoder-decoder blocks, 8 attention heads, a 256-dimensional word representations, a dropout rate of 0.1 and positional feed-forward layers of 1024 inner dimension. Both learning rate schedules and optimization parameters are adopted from the work of \citet{vaswani2017attention}.

\section{Training Protocols}\label{sec: training_protocols}
 All the training protocols employed in this paper are discussed in this section. For all our baseline models, we employ the conventional NMT training protocol as introduced by \citet{sutskever2014sequence}. 
\subsection{Transfer learning}\label{sec:  transfer learning}
To formally define transfer learning we begin by introducing the nomenclature used in the definition. We define our pair  $\mathcal{D} = \{\mathcal{X}, P(X)\}$ as the ``domain'', where the input feature space is denoted by $\mathcal{X}$ and the marginal probability distribution over $X \in \mathcal{X}$ is denoted $P(X)$. Considering the domain $\mathcal{D}$, we define our learning ``task'' as the pair $\mathcal{T}= \{\mathcal{Y}, f(\cdot)\}$ where $\mathcal{Y}$ denotes the target feature space with $f(\cdot)$ the denoting the target predictive function, for example the conditional distribution $P(y|x)$. 

\begin{definition}
Accorded with a source and target domain $\mathcal{D}_{S}$ and $\mathcal{D}_{T}$ respectively, along with their respective tasks $\mathcal{T}_{S}$ and $\mathcal{T}_{T}$. Provided that either $\mathcal{D}_{S} \neq \mathcal{D}_{T}$ or  $\mathcal{T}_{S} \neq \mathcal{T}_{T}$, transfer learning intends to better the predictive function $f_{T}(\cdot)$ by leveraging the learned representations in domain $\mathcal{D}_{S}$ and task $\mathcal{T}_{S}$.
\end{definition}

The transfer learning definition above is conditioned on one of two scenarios: the first being that the source domain $\mathcal{D}_{S}$ and target domain $\mathcal{D}_{T}$ are not equal, $\mathcal{D}_{S} \neq \mathcal{D}_{T}$, which implies that either the source task input feature space and target task input feature space are dissimilar ,$\mathcal{X}_{S} \neq \mathcal{X}_{T}$, or the source and target domain's marginal probability distributions are dissimilar $P_{S}(X) \neq P_{T}(X)$. The second condition states that the source and target tasks are not similar, $\mathcal{T}_{S} \neq \mathcal{T}_{T}$, which suggests that either the two target feature spaces are dissimilar $\mathcal{Y}_{S} \neq \mathcal{Y}_{T}$ or it could be that the two target predictive functions are dissimilar $P(Y_{S}|X_{S}) \neq P(Y_{T}|X_{T})$. In simple terms, transfer learning is an ingenious performance improvement technique of training models by transferring knowledge gained on one task to a related task, especially those with low-resources. \cite{torrey2010transfer}.  
\subsection{Multilingual learning}
Multilingual NMT is a quintessential method of mapping to and from multiple languages and was first suggested by \citet{dong2015multi} with a one-to-many model. Thereafter, this technique was extended to performing many-to-many translations contingent of a task-specific encoder-decoder pair \citep{luong2015multi, firat2016zero}. Afterwards, a single encoder-decoder architecture for performing many-to-many translations was developed \citep{johnson2017google, ha2017effective} by adding a target language specific token at the beginning of each input sequence. In this paper we employ this many-to-many translation technique. 
\begin{definition} 
Given a training tuple $(x_{i}, y_{j})$ where $i,j \in \{1,..,T\}$, the multilingual model's task is to translate source language $i$ to target language $j$. It then follows that the model's objective is to maximize the log-likelihood over all the training sets $D_{i,j}$ appertaining to all the accessible language pairs $\mathcal{S}$:
\begin{equation}
    \max_{\theta}\;\frac{1}{|\mathcal{S}|\cdot|D_{i,j}|}\sum_{(x_{i},y_{j})\in \mathcal{D}_{i,j}, (i,j)\in \mathcal{S}} \log p_{\theta}(y_{j}|x_{i},j^{*}),
\end{equation}
where $j^{*}$ denotes the target language ID.
\end{definition}
The main advantage of a multilingual model is that it leverages the learned representations from individual translation pairs. In other words, the model learns a universal representation space for all the accessible language pairs, which promotes translation of low-resource pairs \citep{firat2016multi, gu2018universal}. 

\subsection{Zero-shot learning}
In addition to facilitating low-resource translations, the universal representation space in multilingual models also facilitates \emph{zero-shot learning}, an extreme variant of transfer learning. First proposed by \citet{johnson2017google} and \citet{ha2017effective}, who demonstrated that multilingual models can translate between untrained language pairs by leverage the universal representations learned during the training process. Adopting the definitions of a task and domain described in Section \ref{sec:  transfer learning}, we give a formal definition of zero-shot learning as follows:
\begin{definition} 
Accorded with source domain $\mathcal{D}_{S} = \{\mathcal{X}_{S}, P(X_{S})\}$ and the respective task $\mathcal{T}_{S} = \{\mathcal{Y}_{S}, P(Y_{S}|X_{S})\}$ where $\mathcal{X_{S}}$ and $\mathcal{Y_{S}}$ denote the input and label feature spaces respectively. The objective of zero-shot learning is to estimate the predictive function or conditional probability distribution $P(Y_{S}|X_{T})$, where $X_{T}$ is the target task input feature space on which the source model has not been trained.
\end{definition}
\section{Related work}\label{sec: related_work}
Notwithstanding the tremendous achievements in the domain of MT, the few publications on MT of African languages bears testament to little growth of this area. The earliest notable work on MT of African languages was done by \citet{wilken2012developing}, where they demonstrated that phrase-based SMT was a promising technique of translating African languages using a English-Setswana language pair. \citet{wolff2014experiments} extended this technique to a E-Z pair, however they employed isiZulu syllables as their source tokens and this modification proved to be efficient as it improved the results by $12.9\%$. \citet{van2014exploring} employed unsupervised word segmentation coupled with phrase-based SMT to translate English to Afrikaans, Northern-Sotho, Tsonga and isiZulu. In their final analysis, the authors found their experiment to be efficient only for Afrikaans and isiZulu datasets, with a SOTA BLEU of $7.9$ for the isiZulu translations of none biblical corpora.\\ \indent \citet{abbott2018towards} adopted NMT to the translation of English to Setswana where they demonstrated that NMT outperformed the previous SMT \cite{wilken2012developing} model by $5.33$ BLEU. The findings of \citet{abbott2018towards} prompted opportunities of extending NMT to other African languages. \citet{martinus2019benchmarking} proposed a benchmark of NMT for translating English to four of South Africa's official languages, namely isiZulu,  Northern Sotho,  Setswana and Afrikaans.\\
\indent Transfer learning has been widely used on non African languages. \citet{zoph2016transfer} investigate the effects of transfer learning in a scenario where the source task (or parent model) is based on a high resource language. This source model is then used to train the target task (or child model). Closely related to the work of \citet{zoph2016transfer} is \citet{nguyen2017transfer} who perform transfer learning on closely related languages. Furthermore, in the work of \citet{nguyen2017transfer}, the authors use a low resource languages as a source model which makes their work closely related to this paper. The main difference being that our work is on African languages.
\begin{table*}[!htb]
\centering
\begin{tabular}{lllll}
\hline
& \textbf{E-S}& \textbf{E-Z} & \textbf{E-X} &  \textbf{X-Z} \\
\hline
\textbf{Sentence count} & 77 500 & 30 253 & 128 342 & 125 098 \\ 
\textbf{Source token count} & 7 588 & 4 649 & 10 657 & 27 144 \\ 
\textbf{Target token count} & 16 408 & 9 424 & 33 947 & 25 465 \\ 
\hline
\end{tabular}
\caption{\label{tab: data_stats}
Summary statistics for our four language pairs. The number of examples in each language is denoted by the sentence count. The number of words in the discrete source languages is denoted by the source sentence token count. Likewise, the word count in the respective target languages is denoted by the target token count.}
\end{table*}

\begin{table}[!htb]
\centering
\begin{tabular}{lllll}
\hline & \textbf{E-S}& \textbf{E-Z} & \textbf{E-X} &  \textbf{X-Z} \\
\hline
\textbf{Train} & 54 250 & 21 177 & 88 192 & 87 570 \\ 
\textbf{Valid} & 11 625 & 4 538 & 20 075 &  18 765 \\ 
 \textbf{Test} & 11 625 & 4 538 & 20 075 & 18 763 \\ 
\hline
\end{tabular}
\caption{\label{tab: data_split} Summary statistics for the number of examples per partition. We split our data into validation, test and set based on a ratio of 3:3:14 respectively. The sum of examples per language pair can be obtained by adding the train, validation and test set examples count, respectively.}
\end{table}
\indent Similarly, zero-shot learning and multilingual modeling have largely been applied on non African languages. For example, both \citet{johnson2017google} and \citet{ha2017effective} applied zero shot learning on languages that are not from Africa. In their work (\citet{johnson2017google} and \citet{ha2017effective}), they show that multilingual systems can perform translation between language pairs that have not been encountered during training. In this work, we seek to leverage these techniques (multilingual and zero-shot learning) on South-Eastern Bantu languages. The applications of zero-shot learning and transfer multilingual learning in this work are closely related to the work of  \citet{lakew2018multilingual} and \citet{johnson2017google}, with the major difference being that this work applies these techniques to low-resource south-eastern Bantu languages.  
\section{Our data}
Our datasets comprise of parallel texts obtained from \citet{orpus64352}, \citet{omniglot64352}, \citet{linguan_64352} and the \citet{wildcoast64352}. As in the case of most machine learning problems, this work included data cleaning, a crucial part of developing machine learning algorithms. The cleaning involved  decomposition of contractions, shortening the sequences, manually verifying some of the randomly selected alignments and dropping all duplicate pairs to curb data leakage. We also decomposed contractions and performed subword level tokenization. The datasets comprise of four language pairs, namely E-S, E-X, E-Z and X-Z. Table \ref{tab: data_stats} gives a summary of the data-set sizes and token count per language pair. We split our data-sets into train, validation and test sets at a ratio of $14:3:3$ respectively, as shown in Table \ref{tab: data_split}.
\section{Experiments}\label{sec: experiments}
\indent For all our training pairs, we train ten transformer models with unique initialization parameters each and take the mean BLEU along with its standard deviation as the final result. We examine the effects of language similarity when performing transfer learning by using E-X and E-S source models to perform on transfer learning on a E-Z task. In this case the similar languages are isiXhosa and isiZulu while on the other hand Shona is the distant language, though still a Bantu language like the other two. We start of by training the base model or source model on a large data-set, for example the E-X pairs. Thereafter we initialize our target model with the source model, whereby the target model is to be trained on a low resourced pair. In other words, instead of starting the target model's training procedure from scratch we leverage the source model's weights to initialize the target model without freezing any of the architecture's layers. \\ \indent The rationale behind this knowledge transfer technique being that in a case where one is faced with few training examples, they can make the most of a prior probability distribution over the model space. Taking the base model (trained on the larger data-set) as the anchor point, the apex of the prior probability distribution belonging to the model space. In setting the base model as the anchor point, the model adapts all the parameters that are useful in both tasks. For example, in the E-S and E-Z tasks, all the source task input language embeddings are adopted to the target task while target language embeddings are modified at training.
\begin{table*}[!htbp]
\centering
\begin{tabular}{lllllll} 
\hline
\multicolumn{2}{l}{\textbf{Model type}}   & \textbf{E-Z}          & \textbf{E-X}          & \textbf{X-Z}          & \textbf{E-S}          & \textbf{E-Z Gain}  \\ 
\hline
\multicolumn{2}{l}{Baseline}  & 8.7 $\pm 0.3$  & 20.9  $\pm 1.4$ &  34.9  $\pm 2.5$   &  16.5 $\pm 0.8$  &    -          \\
\multicolumn{2}{l}{$\mathrm{\text{Multilingual}}_\mathrm{\text{A}}$} &18.6 $\pm 1.0$  & 18.5 $\pm 1.2$     &  30.4 $\pm 1.5$   & -    &    \textbf{\;9.9 $\pm$  \textbf{1.0}} \\
\multicolumn{2}{l}{$\mathrm{\text{Multilingual}}_\mathrm{\text{B}}$} & 6.0 $\pm 0.3$& - & - & 14.3 $\pm 0.2$& \textbf{- 2.7 $\pm \textbf{0.4}$}          \\
\multicolumn{2}{l}{$\mathrm{\text{Multilingual}}_\mathrm{C}$} &14.6$\pm 0.2$  & 18.7 $\pm 0.8$ & -  & -  &    \textbf{ 5.9 $\pm  \textbf{0.3}$}           \\
\multicolumn{2}{l}{Transfer learning $\mathrm{\text{E-X}}_\mathrm{parent}$} &  14.8 $\pm 0.2$    &  -   &  -   &   -  &   \textbf{ 6.1 $\pm \textbf{0.4}$} \\
\multicolumn{2}{l}{Transfer learning $\mathrm{\text{E-S}}_\mathrm{parent}$}&  9.6 $\pm 0.7$   & -    &   -  &   -  & \textbf{\;0.9 $\pm \textbf{0.8}$} \\
\multicolumn{2}{l}{Zero-shot learning}                     &  10.6 $\pm 0.2$  &  18.1 $\pm 1.5$ & 34.0 $\pm 2.4$ & - &      \textbf{ 2.0 $\pm \textbf{0.4}$ }    \\
\hline
\end{tabular}
\caption{\label{tab: baseline_models_bleu} BLEU scores for the baseline, transfer learning, multilingual and zero-shot learning for the language pairs built from English (E), Shona (S), isiXhosa (X), isiZulu (Z). The gains are calculated only for English-to-isiZulu (E-Z), our target pair. Error bars are given by the standard deviations from ten separate re-training of the models in each case. Zero-shot learning applies only to E-Z, built from a  multilingual E-X \& X-Z model.}
\end{table*}
\\ \indent To compare the performance of transfer learning, multilingual leaning and zero-shot learning we train a many-to-many multilingual model that translates E-X, X-Z and E-Z language pairs. The zero-short learning model is trained on E-X and X-Z language pairs.     
\section{Results}\label{sec: results}
Our experimental findings are summarized in Table \ref{tab: baseline_models_bleu}. We obtain the highest baseline BLEU of $34.9 \pm2.5$ on the X-Z model, followed by the E-X model with $20.9 \pm1.4$ BLEU. On the E-S and E-Z baseline models, we obtained $16.5 \pm0.8$ and $8.7 \pm0.3$ BLEU, respectively. As expected, the X-Z language pair had the highest score, mainly owing to the vocabulary overlap between the source language and target language, evidenced by both languages falling under the Nguni sub-class. All three Nguni languages have a lot of vocabulary in common.\\ \indent From the multilingual results, we note that though with some loss in performance, the multilingual model closely resembles the E-X and X-Z baseline model results. The average BLEU loss for the  E-X and X-Z pairs is approximately $2.4 \pm 1.8$ and $4.5 \pm 3$, respectively. This loss is perhaps due to the complexity of learning shared representations across all language pairs. On the other hand, the E-Z pair obtained a BLEU gain of $9.9 \pm1.0$, which is probably a consequence of isiZulu and isiXhosa target languages having significant overlap in vocabulary. As a result, the model's decoder utilizes the learned representations from the target language with more training examples (isiXhosa in this case) to improve the translations with lower training examples (isiZulu in this case).\\ \indent We train the two transfer learning models on a E-Z task, one with the E-X model as the source model and the other with the E-S source model. The transfer learning results show that both source models do improve the BLEU for the E-Z target task, with the greatest improvement coming from the source task that is more closely related to the target task. In our experiments, the E-X task is more closely related to the target task E-Z. The E-S and E-X parent (or source) models had 6.6 and 0.9 BLEU improvements, respectively. These results indicate that the E-X parent model surpassed the E-S model by a significant margin. To be precise, the E-X gain was $5.2 \pm 0.7$ BLEU higher than the E-S model.\\ \indent Compared to the multilingual model, the E-X parent model had $3.8 \pm 1.1$ less BLEU. The E-S parent model on the other hand had $9 \pm 1.3$ less BLEU than multilingual model. These results suggest that with the necessary data available, many to many multilingual modeling is the favourable translation technique. Especially in the case of Southern African languages that have sub-classes with a lot of vocabulary in common. For example, the Nguni, Sotho and Tsonga language sub-classes as shown in Figure \ref{fig: lang_fam_tree}.\\  \indent To perform zero-shot learning, we train our multilingual model on the E-X and X-Z language pairs. The E-X zero-shot preliminary results were $18.1 \pm 1.5$ BLEU, which indicates a $2.8 \pm 2.1$ loss in BLEU. Similarly, this model produced a BLEU of $34.0 \pm 2.4$ on the X-Z language pair, indicating a $0.9 \pm 3.5$ loss in BLEU. On the zero-shot learning task, we recorded a score of $10.6 \pm 0.2$, which is approximately $2.0 \pm 0.4$ BLEU greater than the baseline model. Overall, multilingual learning performs better than transfer learning and zero-shot learning. Second to multilingual learning, transfer learning proved to be an efficient technique only when we employ a parent model with a target language of the same sub-class as the target language in the target task. This is evidenced by the zero-shot learning model obtaining $1.0 \pm 0.7$ BLEU more than the transfer learning model with E-S model as the parent model.   

\section{Conclusion}\label{sec: conclusions}
This work examines the opportunities in leveraging low resource translation techniques in developing translation models for Southern African languages. We focus on English-to-isiZulu (E-Z) as it is the smallest of our corpora with just 30,000 sentence pairs. Using multilingual English-isiXhosa-isiZulu learning we achieve a BLEU score for English-to-isiZulu  of 18.6 $\pm 1.0$, more than doubling the previous  state-of-the-art and yielding significant gains (9.9 in BLEU score) over the baseline English-to-isiZulu transformer model.  Multilingual learning for this dataset outperforms both transfer learning and zero-shot learning, though both of these techniques are better than the baseline mode, with BLEU score gains of 6.1 and 2.0 respectively. \\ \indent We further found that transfer learning is a highly effective technique for training low resource translation models for closely related South-Eastern Bantu languages. Using the English-to-isiXhosa baseline model, transfer learning to isiZulu had a BLEU score gain of 6.1 while using the English-to-Shona baseline model for transfer learning yielded no statistical significant gain at all ($0.9 \pm 0.8$). Since isiXhosa is similar to isiZulu while Shona is quite different, this illustrates the performance gains that can be achieved by exploiting language inter-relationships with transfer learning, a conclusion further emphasised by the fact that zero-shot learning, in which no English-to-isiZulu training data was available, outperformed transfer learning using the English-to-Shona baseline model.

\section{Acknowledgements}
Our greatest appreciation goes to Jade Abbott and Herman Kamper for discussions and the African Institute for Mathematical Sciences (AIMS) and the National Research Foundation of South Africa, for supporting this work with a research grant. We further express our sincere gratitude to the South African Centre for High-Performance Computing (CHPC) for CPU and GPU cluster access.

\bibliographystyle{acl_natbib}
\bibliography{IEEEabrv,eacl2021}

\appendix

\end{document}